\documentclass[conference]{IEEEtran}
\IEEEoverridecommandlockouts
\usepackage{cite}
\usepackage{amsmath,amssymb,amsfonts}
\usepackage{graphicx}
\usepackage{textcomp}
\usepackage{xcolor}
\def\BibTeX{{\rm B\kern-.05em{\sc i\kern-.025em b}\kern-.08em
    T\kern-.1667em\lower.7ex\hbox{E}\kern-.125emX}}
    
\newcommand{\oneLine}[1]{\rotatebox[origin=c]{90}{$\text{#1}$}}
\newcommand{\twoLines}[2]{\rotatebox[origin=c]{90}{$\text{#1}\atop\text{#2}$}}

\newcommand{\refeq}[1]{Eq.~\eqref{#1}}
\newcommand{\reffig}[1]{Fig.~\ref{#1}}

\usepackage{listings}
\usepackage{ifxetex}
\usepackage{fancyhdr}
\usepackage{hyperref}
\usepackage{graphicx}
\usepackage{wrapfig}
\usepackage{amsmath}
\usepackage{amsfonts}
\usepackage{amssymb}
\usepackage{listings}
\usepackage{setspace}
\usepackage{wrapfig}
\usepackage{tablefootnote}
\usepackage{multirow}
\usepackage{booktabs}
\usepackage[section]{placeins}
\usepackage{color}
\usepackage{amsfonts}
\usepackage{ulem}
\usepackage{algorithm}
\usepackage{algorithmicx}
\usepackage{algpseudocode}
\usepackage{graphicx}
\usepackage{float}
\usepackage{dsfont}
\usepackage{cleveref}

\newcommand{\mat}[1]{\mathbf{#1}}
\newcommand{\set}[1]{\mathcal{#1}}
\newcommand{\pnorm}[1]{\lVert{#1}\rVert}

\DeclareMathOperator*{\dist}{{d}}

\DeclareMathOperator*{\SetSymMat}{\mathcal{S}}

\DeclareMathOperator*{\diag}{diag}
\DeclareMathOperator*{\sign}{sign}

\newcommand{\N}{\ensuremath{\mathcal{N}}} 
\newcommand{\I}{\mat{\mathbb{I}}}

\newcommand{\RN}{\mathbb{R}}

\DeclareMathOperator*{\densitythreshold}{\ensuremath{\delta}}

\newcommand{\densityestimator}{\ensuremath{\hat{p}}}

\newcommand{\densityestimatorgmm}{\densityestimator_{\text{GMM}}}

\DeclareMathOperator*{\regularization}{\ensuremath{{\theta}}}
\DeclareMathOperator*{\prototype}{{p}}
\newcommand{\x}{\ensuremath{\vec{x}}}
\newcommand{\z}{\ensuremath{\vec{z}}}
\newcommand{\y}{\ensuremath{y}}
\newcommand{\xorig}{\ensuremath{\vec{x}_{\text{orig}}}}
\newcommand{\yorig}{\ensuremath{y_{\text{orig}}}}

\newcommand{\sety}{\ensuremath{\set{Y}}}
\newcommand{\xcf}{\ensuremath{\vec{x}'}}
\newcommand{\ycf}{\ensuremath{y'}}
\newcommand{\deltacf}{\ensuremath{\vec{\delta}}}
\newcommand{\basisvalues}{\ensuremath{\vec{b}}}
\newcommand{\w}{\ensuremath{\vec{w}}}
\newcommand{\q}{\ensuremath{\vec{q}}}
\newcommand{\setI}{\set{I}}
\newcommand{\dimsym}{d}
\newcommand{\classifier}{\ensuremath{h}}

\newcommand{\protolabel}{\ensuremath{o}}
\DeclareMathOperator*{\distmat}{{\mat{\Omega}}}

\newboolean{arxiv}
\setboolean{arxiv}{true}     
    
\begin{document}

\title{Efficient computation of contrastive explanations\\
\thanks{We gratefully acknowledge funding from the VW-Foundation for the project \textit{IMPACT} funded in the frame of the funding line \textit{AI and its Implications for Future Society}.}}

\author{\IEEEauthorblockN{1\textsuperscript{st} Andr\'e Artelt}
\IEEEauthorblockA{\textit{CITEC - Cognitive Interaction Technology} \\
\textit{Bielefeld University}\\
Bielefeld, Germany \\
aartelt@techfak.uni-bielefeld.de}
\and
\IEEEauthorblockN{2\textsuperscript{nd} Barbara Hammer}
\IEEEauthorblockA{\textit{CITEC - Cognitive Interaction Technology} \\
\textit{Bielefeld University}\\
Bielefeld, Germany \\
bhammer@techfak.uni-bielefeld.de}
}

\maketitle

\begin{abstract}
With the increasing deployment of machine learning systems in practice, transparency and explainability have become serious issues.
Contrastive explanations are considered to be useful and intuitive, in particular when it comes to explaining decisions to lay people, since they mimic the way in which humans explain. Yet, so far, comparably little research has addressed computationally feasible technologies, which allow guarantees on uniqueness and optimality of the explanation and which enable an easy 
incorporation of additional constraints.
Here, we will focus on specific types of models rather than black-box technologies.
We study the relation of contrastive and counterfactual explanations and propose mathematical formalizations as well as a 2-phase algorithm for efficiently computing (plausible) pertinent positives of many standard machine learning models.

\end{abstract}


\section{Introduction}
The increasing deployment of machine learning (ML) systems in practice led to an increased interest in explainability and transparency. In particular, ``prominent failures'' of ML systems like predictive policing~\cite{predictivepolicing}, loan approval~\cite{creditscoresunfair} and face recognition~\cite{failurefacerecognition}, highlighted the importance of transparency and explainability of ML systems. In addition, the need for explainability was also recognized by policy makers which resulted in a ``right to an explanation'' in the EU’s ``General Data Protection Right'' (GDPR)~\cite{gdpr}.
The crucial problem with regard to these demands is the definition and the type of explanations - there exist many different kinds of explanations~\cite{molnar2019,explainingblackboxmodelssurvey,explainingexplanations,explainableartificialintelligence,surveyxai} but it is still not clear how to properly formalize an explanation~\cite{molnar2019, doshi2017towards}.

One family of explanations are example-based explanations~\cite{casebasedreasoning} which are considered to be particularly well suited for lay people, since they allow the inspection of explanations by looking at example data, including  the possibility of domain-specific representations of data~\cite{molnar2019}. Counterfactual explanations~\cite{counterfactualwachter} and contrastive explanations constitute instantiations of example-based explanations~\cite{contrastiveexplanationswithpertinentnegatives, molnar2019, accountabilityAI}; these will be the focus in this work.

Following the common definition/intuition of a contrastive explanation~\cite{molnar2019, accountabilityAI} (in the context of~\cite{contrastiveexplanationswithpertinentnegatives}), a contrastive explanation consists of two parts:
\begin{itemize}
\item A \textit{pertinent positive} specifies a minimal and interpretable set of features that must be present for obtaining the same prediction as the complete sample does. Meaning that we are looking for a subset of features 
such that the resulting sample has the same prediction as the original sample.

\item A \textit{pertinent negative} specifies a
set of features, which \textit{must not} be present to provide the prediction, i.e.\ it is  contrastive, since it relates to elements representative of a different class which are absent; expressed in different words, it refers to a small and interpretable perturbation of the original sample that would lead to a different prediction than the original sample.
\end{itemize}
Together, a pertinent negative and pertinent positive form a contrastive explanation.

For an example, consider the application of a loan approval system. Imagine that the system rejects a loan application and we now have to explain its decision. A possible contrastive explanation (consisting of a pertinent negative and a pertinent positive) might be: \textit{The loan application was rejected because the pay back of the last loan was delayed, the applicant has a second credit card and because the monthly income is not above a minimum specific threshold, required for acceptance of the loan.} The first two arguments/reasons can be considered as a pertinent positive and the last reason as a pertinent negative.
Note that, if more than two classes are present, pertinent negatives always contrast the present class to one specified alternative class.

\paragraph*{Related work}
There does exist extensive work and experimental evidence, which highlights that explanations provided by people 
are often contrastive in nature  \cite{DBLP:journals/corr/Miller17a}:
rather than explaining reasons for an observed event $p$, people often focus on reasons for observing $p$ rather than another specific event $q$. The question of how to compute contrastive explanations for technical systems, constitutes an issue, though. In causal models, contrastive arguments of factors, which explain an appearance of $p$ rather than $q$, can be  based on according triangulations within the logical relations
\cite{Lipton1991-LIPCEA}. For black box models including deep networks,
there exists some  work how to compute 
contrastive explanations in practice~\cite{contrastiveexplanationswithpertinentnegatives, contrastiveexplanationsmodelagnostic, contrastiveexplanationslocalfoiltrees}.
More specifically,
the authors of~\cite{contrastiveexplanationswithpertinentnegatives} propose an algorithm called ``contrastive explanation method (CEM)'' that computes a contrastive explanation of a differentiable model such as a Deep Neural Network. The method computes a pertinent positive and a pertinent negative by solving strongly regularized cost functions by using a projected fast iterative shrinkage-thresholding (FISTA) algorithm. A part of the regularizations consists of an autoencoder ensuring that the solution is plausible. While this approach might be well suited for Deep Neural Networks, it might be less suited for standard ML models, where the regularization is not clear, and an autoencoder is not easily  available, e.g.\ because the training set is too small. Furthermore, there do not exist  theoretical guarantees of the result, in particular the sensitivity of the provided explanations   with respect to the chosen regularization can be high.

In subsequent work~\cite{contrastiveexplanationsmodelagnostic}, the authors extend CEM towards the model agnostic contrastive explanation method (MACEM) for computing contrastive explanations of an arbitrary (not necessarily differentiable) model. The modelling approach is somewhat similar to the one in~\cite{contrastiveexplanationswithpertinentnegatives}. MACEM uses FISTA and estimates the gradient in case of a fully black-box (not-differentiable) model. Furthermore, the authors also propose how to model categorical features. 

The authors of~\cite{contrastiveexplanationslocalfoiltrees} address model agnostic contrastive explanations, which are obtained based on locally trained decision trees which serve as a  local surrogate of the observed model. Since this method needs to sample training points around a given data point, it is sensitive to the curse of dimensionality.

Most of the methods for computing contrastive explanations are somewhat model agnostic or are suitable for a ``broader'' class of models. 
As a consequence, it is not easily possible to provide guarantees on important properties such as uniqueness of the explanation, since no assumptions on the type of
model are made. Further, the involved optimization technologies might be computationally demanding, and they often rely on  iterative numeric methods such as general gradient-based optimization technologies.
Here, we are interested in the question, how to efficiently compute contrastive explanations for specific models, which are popular in machine learning.
For specific models, a general method might not be the most efficient one and specific formulations might provide particularly efficient alternatives, for which additional guarantees such as convexity and uniqueness hold. In this work we study how to exploit model specific structures for efficiently computing contrastive explanations of several standard ML models. To the best of our knowledge, this is the first work to address the question how to efficiently compute such model-specific contrastive explanations.

\paragraph*{Our contributions}
We make several contributions in this work:
\begin{enumerate}
\item In section~\ref{sec:pertinentnegatives} we address a conceptual issue, and we study how pertinent negatives are related to counterfactual explanations as discussed e.g.\ in \cite{face}. We reduce the problem of computing a pertinent negative to the problem of computing a counterfactual explanation. For the latter,  model-specific optimization schemes have recently been proposed in the work~\cite{counterfactualsurvey}.
\item In section~\ref{sec:pertinentpositives} we conceptualize  computing pertinent positives and we propose a 2-phase algorithm for computing ``high-quality'' pertinent positives. In section~\ref{sec:pp:modelspecificprograms} we develop mathematical programs (often even convex programs) for efficiently computing pertinent positives of many different standard ML models like linear/quadratic classifiers and learning vector quantization models.
We also study how to compute plausible pertinent positives, and we also study special cases in which we can efficiently compute globally optimal pertinent positives.
\item We empirically evaluate our proposed methods in section~\ref{sec:pp:experiments}.
For most settings, we obtain unique explanations.
\end{enumerate}
\ifthenelse{\boolean{arxiv}}
{Due to space constraints and for the purpose of better readability, we include all proofs and derivations to the appendix (section~\ref{sec:appendix}).}
{Due to space constraints, all proofs and derivations can be found in an extended version on arXiv\footnote{\url{https://arxiv.org/abs/2010.02647}}.}

\section{Pertinent negatives as counterfactuals}\label{sec:pertinentnegatives}
A pertinent negative, as described in~\cite{contrastiveexplanationswithpertinentnegatives}, specifies a ``small and interpretable'' perturbation $\deltacf$ of the original sample $\xorig$ that leads to a different prediction $\ycf\neq\yorig$, i.e.\ it contrasts the current output $\yorig$ to another class $\ycf$.
If we consider a small $1$-norm as ``small and interpretable'', we can phrase the computation of a pertinent negative as the following optimization problem:
\begin{subequations}\label{eq:opt:pertinentnegative}
\begin{align}
&\underset{\deltacf\,\in\,\RN^{\dimsym}}{\min}\, \pnorm{\deltacf}_{1} \quad\quad \text{s.t. } \classifier(\xorig + \deltacf) = \ycf \neq \yorig  \label{eq:opt:pertinentnegative:constraint}
\end{align}
\end{subequations}
where $\classifier:\RN^{\dimsym}\to\sety$ denotes the classifier whose prediction we want to explain.
Here, the 1-norm accounts not only for a small change, but also sparsity as regards the number of features, which are changed.

The constrained optimization problem for computing a counterfactual explanation~\cite{counterfactualwachter} as proposed by~\cite{counterfactualsurvey} is given as:
\begin{subequations}\label{eq:opt:counterfactual}
\begin{align}
& \underset{\xcf\,\in\,\RN^d}{\min}\, \regularization(\xorig, \xcf) \quad\quad \text{s.t. } \classifier(\xcf) = \ycf \label{eq:opt:counterfactual:constraint}
\end{align}
\end{subequations}
where $\regularization(\cdot)$ denotes a regularization (e.g. $1$-norm), $\xcf$ denotes the counterfactual and $\ycf\neq\yorig$ the requested target label.

We can turn \refeq{eq:opt:pertinentnegative} into \refeq{eq:opt:counterfactual} by setting $\xcf = \xorig + \deltacf$ and choosing $\regularization(\xorig, \xcf)=\pnorm{\xorig - \xcf}_{1}$. The appealing consequence of this is that we can reduce the problem of computing a pertinent negative to computing a counterfactual explanation for which several efficient methods already exists~\cite{counterfactualwachter,counterfactualsurvey, counterfactualguidedbyprototypes,counterfactualsgeneticalgorithm}. The work \cite{counterfactualsurvey}, in particular, proposes convex formulations of the problem for a number of important ML models. The work
\cite{plausiblecounterfactuals} enriches this framework with efficient approximations of 
how to compute plausible counterfactuals with a guaranteed likelihood  value, in order to distinguish
those from adversarial examples, which correspond to artificial signals in particular for high dimensional data \cite{43405}.

Note that the computation of pertinent negatives as counterfactual explanations perfectly fits the intuition of contrasting the given prediction $\yorig$ against some other (predefined) prediction $\ycf$ as discussed in the introduction of this work.


\section{Pertinent positives}\label{sec:pertinentpositives}
\subsection{Modelling}\label{sec:pp:modelling}
In order to model the intuition of a pertinent positive, as described in~\cite{contrastiveexplanationswithpertinentnegatives}, we have to consider several aspects:
\begin{itemize}
    \item We want to ``turn off'' as many features as possible.
    \item For ``turned on'' features, the difference to the original feature values should be as small as possible.
    \item The pertinent positive must be still classified as $\yorig$.
\end{itemize}
We denote the final pertinent positive\footnote{It is debatable (and of course highly dependent on the use-case) whether the data point $\xcf$ or the perturbation $\deltacf$ is presented as the ``explanation'' to the user.} $\xcf$ as:
\begin{equation}\label{eq:pertinentpositive}
\xcf = \xorig - \deltacf
\end{equation}
where $\deltacf$ denotes the perturbation (changes) that give rise to the pertinent positive.
In order to improve readability of the subsequent formulas, we will sometimes substitute \refeq{eq:pertinentpositive} and optimize over $\xcf$ instead of $\deltacf$ - we mean by this an optimization over $\deltacf$ which implies $\xcf$.

Mathematically, we formalize the fact that a feature is ``turned off'' by its value being identical to zero.
Like the authors of~\cite{contrastiveexplanationsmodelagnostic} did, we can always subtract a constant $\basisvalues$ from the original sample $\xorig$ to allow non-zero default values - i.e. $\basisvalues$ would denote the feature wise base/default values at which we consider a particular feature to be ``turned off'', in the sense that a feature does not deviate ``much'' from the default value (e.g.\ the expected value or a statistically robust estimation thereof). In the following, we assume $\basisvalues=0$ for simplicity.

Considering all these aspects yields the following multi-objective optimization problem:
\begin{subequations}\label{eq:pertinentpositive:modelling}
\begin{align}
    & \underset{\deltacf\,\in\,\RN^{\dimsym}}{\min}\; \Big|\left[\xorig - \xcf\right]_{\setI}\Big| \quad \text{where } \xcf = \xorig - \deltacf \label{eq:pertinentpositive:modelling:closeness}\\
    & \underset{\setI}{\min}\; \left|\setI\right| \label{eq:pertinentpositive:modelling:sparsity}\\
    & \text{s.t.} \quad \classifier(\xcf) = \yorig
\end{align}
\end{subequations}
where $[\cdot]_{\setI}$ denotes the selection operator on the set $\setI$, whereby $\setI$ denotes the set of all ``turned on'' features.\footnote{The selection operator returns a vector whereby it only selects a subset of indices from the original vector as specified in the set $\setI$.} $\setI$ is defined as follows:
\begin{equation}\label{eq:setI}
\setI = \Big\{i:\,\big|(\xcf)_i\big|>\epsilon\Big\}
\end{equation}
where $\epsilon\in\RN_{+}$ denotes a tolerance threshold at which we consider a feature ``to be turned on'' - e.g. a strict choice would be $\epsilon=0$.

The optimization problem \refeq{eq:pertinentpositive:modelling} is difficult, since it is highly non-convex, it contains a discrete variable, and the two objectives \refeq{eq:pertinentpositive:modelling:closeness} and \refeq{eq:pertinentpositive:modelling:sparsity} are in parts contradictory. Therefore, we propose a relaxation in the subsequent section. This relaxation allows us to efficiently compute pertinent positives of many standard ML models (we will turn this relaxation into a convex relaxation for many standard ML models) - we empirically evaluate our proposed relaxation in the experiments (see section~\ref{sec:pp:experiments}).

\subsection{Relaxation by a 2-phase algorithm}\label{sec:pp:2phasealgo}
For computing a pertinent positive \refeq{eq:pertinentpositive:modelling}, we have to ensure sparsity and closeness to the original sample. We propose to approximately solve \refeq{eq:pertinentpositive:modelling} by a 2-phase algorithm where we separate the computational goals, sparsity and closeness, in two phases.

\subsubsection{Sparsity}
In order to achieve sparsity of the pertinent positive, we propose the following optimization for ensuring a sparse pertinent positive:
\begin{subequations}\label{eq:opt:pertinentpositive}
\begin{align}
&\underset{\deltacf\,\in\,\RN^d}{\min}\, \pnorm{\xorig - \deltacf}_{1} \label{eq:opt:pertinentpositive:objective} \\
& \text{s.t.} \quad \classifier(\xorig - \deltacf) = \yorig \label{eq:opt:pertinentpositive:constraint}
\end{align} 
\end{subequations}

Although the optimization problem \refeq{eq:opt:pertinentpositive} looks similar to the one proposed in~\cite{contrastiveexplanationswithpertinentnegatives,contrastiveexplanationsmodelagnostic}, a crucial difference is that \refeq{eq:opt:pertinentpositive} is a constrained optimization problem with a convex objective - this is what allows us (see section~\ref{sec:pp:modelspecificprograms}) to derive convex programs for computing pertinent positives of many standard ML models. Sparsity is here enforced by the $1$-norm, instead of the $0$-norm.
Furthermore, our formulation \refeq{eq:opt:pertinentpositive} allows to easily add additional constraints like box constraints or ``freezing'' some features, for meeting domain specific requirements (e.g. plausibility).
Another consequence of our modelling is that we do not need any hyperparameters - note that the formulation in~\cite{contrastiveexplanationswithpertinentnegatives} uses several hyperparameters that have to be chosen. Since our formulation comes without any hyperparameters, the computation is easier. More importantly, by making use of convex optimization we can provide theoretical guarantees  such as  uniqueness or an exact statement of existence or non-existence of a solution.

\subsubsection{Closeness}\label{sec:pp:2phasealgo:closeness}
By solving the optimization problem \refeq{eq:opt:pertinentpositive} we obtain a sparse pertinent positive. As already discussed, while sparsity is in alignment with the intuition of a pertinent positive, it can happen that many features will be shrunken towards zero and thus be far away from the original features values - we will empirically observe this behavior in the experiments (section~\ref{sec:pp:experiments}) -, which contradicts the intuition of a pertinent positive. Therefore, we propose a second optimization step, enforcing closeness for the values, which are kept.

Also note that it can happen that the optimal solution of \refeq{eq:opt:pertinentpositive} is the zero vector $\vec{0}$; this holds  if  the zero vector is classified as the same class as the original sample - i.e. $\classifier(\vec{0})=\yorig$. In this case, all features would be ``turned off'.' We argue that in this case a pertinent positive might not make much sense because such an explanation would not be very informative for the user, and it is unclear how to break symmetries about which features are relevant in this case. Note that this might change when considering plausible pertinent positives instead of sparsest pertinent positives - see section~\ref{sec:pp:plausibility}.
We propose to reduce an explanation to the pertinent negative part, in this case, or to add additional semantic information, which indicates which features are relevant. As an example, one could avoid this issue by fixing some features to their original values or introducing box constraints that prevent a certain number of features of being ``turned off''. Such kind of constraints easily fit into the proposed optimization problems \refeq{eq:opt:pertinentpositive} and \refeq{eq:opt:improve_pp} and do not change the computational complexity of the problems.



Provided the first phase of the algorithm yields a reasonable and non-trivial solution $\{j:|\: |(\xcf)_i| \leq \epsilon\}$ for features which can be turned off, where
$\xcf$ is the solution from \refeq{eq:opt:pertinentpositive},
we propose a second phase, where we minimize the distance of the remaining features to the original values, as follows: 
\begin{subequations}\label{eq:opt:improve_pp}
\begin{align}
& \underset{\xcf\,\in\,\RN^d}{\min} \sum_{i\,\in\,\setI} \big|(\xcf)_i - (\xorig)_i\big| \label{eq:opt:improve_pp:objective} \\
& \text{s.t.}\quad\classifier(\xcf) = \yorig \label{eq:opt:improve_pp:constraint}\\
&\quad\quad |(\xcf)_i| \leq \epsilon \quad\forall\,i\not\in\setI \label{eq:opt:improve_pp:constraint:fixturnedofffeatures}
\end{align}
\end{subequations}
The final 2-phase algorithm is described as pseudo code in Algorithm~\ref{algo:2phase} and is empirically evaluated in section~\ref{sec:pp:experiments} (Experiments).
\begin{algorithm}[!tb]
\caption{Computation of a pertinent positive}\label{algo:2phase}
\begin{algorithmic}[1]
	\Require A labeled sample $(\xorig,\yorig)$
	\Ensure A pertinent positive $\xcf$
	\State Compute a pertinent positive $\xcf$ by solving \refeq{eq:opt:pertinentpositive}
	\State Try to improve $\xcf$ by solving \refeq{eq:opt:improve_pp}
\end{algorithmic}
\end{algorithm}
Interestingly, this two-step algorithm can be instantiated as efficient convex problems for many popular machine learning models, as we will show in the following.

\subsection{Model specific programs}\label{sec:pp:modelspecificprograms}
In the subsequent sections we study how the optimization problem \refeq{eq:opt:pertinentpositive} evolves for different standard ML models - in particular we reduce \refeq{eq:opt:pertinentpositive} to convex or ``nearly convex'' programs.
Because the objectives \refeq{eq:opt:improve_pp:objective} and \refeq{eq:opt:pertinentpositive:objective} are both convex and independent of the model $\classifier$, it is sufficient to work on \refeq{eq:opt:pertinentpositive} only - if we can turn \refeq{eq:opt:pertinentpositive} into a convex program (meaning we have to turn the constraint \refeq{eq:opt:improve_pp:constraint} into a convex one), then the same holds for \refeq{eq:opt:improve_pp}.

\subsubsection{Linear models}\label{sec:pp:linearmodels}
A linear classifier $\classifier: \RN^{\dimsym} \to \sety$ can be written as follows:
\begin{equation}\label{eq:linearmodel}
\classifier(\x) = \sign(\w^\top\x + b)
\end{equation}
where we restrict our-self to a binary classifier - however, the idea (and everything that follows) can be generalized to multi-class problems.
Popular instances of linear models are logistic regression, linear discriminant analysis (LDA) and linear support vector machine (linear-SVM).

Assuming $\sety=\{-1, 1\}$, we can rewrite the constraint \refeq{eq:opt:pertinentpositive:constraint} as follows:
\begin{equation}\label{eq:linearmodel:pertinentpositive:constraint}
\yorig \w^\top\deltacf - c + \epsilon \leq 0
\end{equation}
where $\epsilon$ denotes a small positive constant that ensures that the set of feasible solutions is closed (strict vs. non-strict inequality) and
\begin{equation}
c = \yorig \w^\top\xorig + \yorig b
\end{equation}
Note that \refeq{eq:linearmodel:pertinentpositive:constraint} is linear in $\deltacf$ and because the objectives \refeq{eq:opt:pertinentpositive:objective} and \refeq{eq:opt:improve_pp:objective} are linear, the optimization problems become linear programs which can be solved efficiently~\cite{Boyd2004}. \ifthenelse{\boolean{arxiv}}{The derivation of \refeq{eq:linearmodel:pertinentpositive:constraint} can be found in appendix~\ref{appendix:pp:linearmodels}.}{}

\subsubsection{Quadratic models}\label{sec:pp:quadraticmodels}
A quadratic classifier $\classifier: \RN^{\dimsym} \to \sety$ can be written as follows:
\begin{equation}\label{eq:quadraticmodel}
\classifier(\x) = \sign(\x^\top\mat{Q}\x + \vec{q}^\top\x + c)
\end{equation}
where $\mat{Q}\in\SetSymMat^{\dimsym}$ and again we restrict our-self to a binary classifier - again, the idea (and everything that follows) can be generalized to multi-class problems.
Popular instances of quadratic models are quadratic discriminant analysis (QDA) and Gaussian Naive Bayes.

Again, if we assume $\sety=\{-1,1\}$, we can rewrite the constraint \refeq{eq:opt:pertinentpositive:constraint} as the following quadratic constraint:
\begin{equation}\label{eq:quadraticmodel:pertinentpositive:constraint}
\deltacf^\top\mat{\tilde{Q}}\deltacf + \deltacf^\top\z + c' + \epsilon \leq 0
\end{equation}
where
\begin{equation}
\begin{split}
&\mat{\tilde{Q}} = -\yorig\mat{Q} \quad\quad \z = 2\yorig\xorig^\top\mat{Q} \\
&c' = -\yorig\Big(\xorig^\top\mat{Q}\xorig + \q^\top\xorig + c\Big)
\end{split}
\end{equation}
Since all we know about $\mat{\tilde{Q}}$ is that it is symmetric, \refeq{eq:quadraticmodel:pertinentpositive:constraint} is in general non-convex. Solving non-convex quadratic programs is known to be NP-hard~\cite{Boyd2004,qcqp}. However, we can rewrite \refeq{eq:quadraticmodel:pertinentpositive:constraint} as a difference of two convex functions\footnote{Every symmetric matrix can be written as the difference of two s.psd. matrices.} and thus turn the whole program into a special instance of a difference of convex programming (DC) for which efficient approximation solvers exist\ifthenelse{\boolean{arxiv}}{ - more details can be found in appendix~\ref{appendix:pp:quadraticmodels}.}{.}

\subsubsection{Learning vector quantization models}\label{sec:pp:lvqmodels}
In learning vector quantization (LVQ) models~\cite{lvqreview} we compute a set of labeled prototypes $\{(\vec{\prototype}_i, \protolabel_i)\}$ from a training data set of labeled real-valued vectors - we refer to the $i$-th prototype as $\vec{\prototype}_i$ and the corresponding label as $\protolabel_i$.
A new data point is classified according to the winner-takes-it-all scheme:
\begin{equation}\label{eq:lvqmodel}
\begin{split}
& \classifier: \x \mapsto \protolabel_i \quad\quad \text{s.t. } \vec{\prototype}_i = \underset{\vec{\prototype}_j}{\arg\min}\;{\dist}(\x, \vec{\prototype}_j)
\end{split}
\end{equation}
where $\dist(\cdot)$ denotes a distance function. In vanilla LVQ, this is chosen globally as the squared Euclidean distance\\
\mbox{$\dist(\x, \vec{\prototype}_j) = (\x - \vec{\prototype}_j)^\top \I (\x - \vec{\prototype}_j)$}.
There exist extensions to  a global quadratic form \mbox{$\dist(\x, \vec{\prototype}_j) = (\x - \vec{\prototype}_j)^\top{\distmat}(\x - \vec{\prototype}_j)$} with $\distmat\in\SetSymMat_{+}^{d}$, referred to as matrix-LVQ (GMLVQ)~\cite{gmlvq}, or a  prototype specific quadratic form \mbox{$\dist(\x, \vec{\prototype}_j) = (\x - \vec{\prototype}_j)^\top{\distmat}_{j}(\x - \vec{\prototype}_j)$} with $\distmat_j\in\SetSymMat_{+}^{d}$, referred to as local-matrix LVQ (LGMLVQ)~\cite{mrslvq}.

Similar to the algorithm for computing counterfactual explanations of LVQ models~\cite{counterfactuallvq}, the idea is to use a Divide-Conquer approach for computing a pertinent positive of a LVQ model \refeq{eq:lvqmodel}. Because the LVQ model outputs the label of the closest prototype, we know that in order to get a specific prediction $\y=\yorig$, the closest prototype must be one the prototypes labeled as $\yorig$. Therefore, we simply try all possible prototypes (labeled as $\yorig$) and select the one that leads to the smallest objective \refeq{eq:opt:pertinentpositive:objective}. For every suitable prototype $\vec{\prototype}_i$, we can rewrite the constraint \refeq{eq:opt:pertinentpositive:constraint} as follows:
\begin{equation}\label{eq:pp:lvq:constraint}
\deltacf^\top\mat{A}_{ij}\deltacf + \deltacf^\top q_{ij} + c_{ij} + \epsilon \leq 0 \quad \forall\,j: \,\protolabel_j\neq\yorig
\end{equation}
where
\begin{equation}
\begin{split}
&\mat{A}_{ij} = {\distmat}_i - {\distmat}_j \quad \vec{q}_{ij}=2{\distmat}_j \left(\xorig - \vec{\prototype}_j\right) - 2{\distmat}_i \left(\xorig - \vec{\prototype}_i\right)\\
&c_{ij} = \left(\xorig - \vec{\prototype}_i\right)^\top{\distmat}_i\left(\xorig - \vec{\prototype}_i\right) -\\&\quad\quad \left(\xorig - \vec{\prototype}_j\right)^\top{\distmat}_j\left(\xorig - \vec{\prototype}_j\right)
\end{split}
\end{equation}
In case of GMLVQ, the constraints \refeq{eq:pp:lvq:constraint} become linear while in the case of LGMLVQ the constraints ~\refeq{eq:pp:lvq:constraint} become quadratic (but potentially non-convex). Because the objective \refeq{eq:opt:pertinentpositive:objective} is linear, \refeq{eq:opt:pertinentpositive} becomes a linear program in case of GMLVQ and a (non-convex) quadratic program in case of LGMLVQ. Again, while linear programs can be solved very efficiently~\cite{Boyd2004}, (non-convex) quadratic programs can not (unless they turn out to be convex quadratic programs)~\cite{Boyd2004,qcqp}. Like in the case of quadratic classifiers, we can easily rewrite the constraint \refeq{eq:pp:lvq:constraint} as a difference of convex functions and then turn the whole program into a special instance of a DC for which good approximation solvers exist~\cite{qcqp}\ifthenelse{\boolean{arxiv}}{ - more details can be found in appendix~\ref{appendix:pp:lvqmodels}.}{.}

The resulting algorithm is summarized in Algorithm~\ref{algo:pp:lvq}.
\begin{algorithm}[!tb]
\caption{Computing a pertinent positive of a LVQ model}\label{algo:pp:lvq}
\begin{algorithmic}[1]
 \Require Labeled sample $(\xorig,\yorig)$ and the LVQ model
 \Ensure Pertinent positive $\xcf$
 \State $\xcf = \vec{0}$ \Comment{Initialize dummy solution}
 \State $z = \infty$
 \For{$\vec{\prototype}_i$ with $\protolabel_i=\yorig$} \Comment{Try each prototype with a suitable label}
 	\State Solving \refeq{eq:opt:pertinentpositive} (substitute \refeq{eq:opt:pertinentpositive:constraint} with \refeq{eq:pp:lvq:constraint}) yields a pertinent positive $\xcf_{*}$
 	\If{$\pnorm{\xcf_{*}}_{1} < z$} \Comment{Keep this pertinent positive if it is sparser than the currently ``best'' pertinent positive}
 		\State $z=\pnorm{\xcf_{*}}_{1}$
 		\State $\xcf = \xcf_{*}$
 	\EndIf
 \EndFor
\end{algorithmic}
\end{algorithm}
Note that the \texttt{for} loop in Algorithm~\ref{algo:pp:lvq} can be easily parallelized because it does not matter when we compute the minimum.

\subsection{Exact solutions for special cases}\label{sec:pp:exactsolutions}
An alternative to the original modelling of a pertinent positive~\refeq{eq:pertinentpositive:modelling}, is a stricter variant that do not allow any deviations for ``turned on'' features.
Instead of~\refeq{eq:pertinentpositive:modelling}, we propose the following similar modelling for computing a pertinent positive:
\begin{subequations}\label{eq:pertinentpositive:modelling:strict}
\begin{align}
    &\underset{\setI}{\min}\,|\setI| \label{eq:pertinentpositive:modelling:strict:objective} \\
    &\text{s.t. } \classifier\Big(\left[\xorig\right]_{\setI}\Big)=\yorig \label{eq:pertinentpositive:modelling:strict:constraint} 
\end{align}
\end{subequations}
Note that in contrast to~\refeq{eq:pertinentpositive:modelling}, we require that all selected (``turned on'') features are equal to their original values.

Although~\refeq{eq:pertinentpositive:modelling:strict} is similar to~\refeq{eq:pertinentpositive:modelling}, both modellings are not equivalent - however, a feasible solution of~\refeq{eq:pertinentpositive:modelling:strict} is also feasible under~\refeq{eq:pertinentpositive:modelling}.

In general,~\refeq{eq:pertinentpositive:modelling:strict} can be interpreted as a feature selection problem which usually are computational difficult to solve exactly. However, in some cases (special instances of the classifier $\classifier$) we can globally solve~\refeq{eq:pertinentpositive:modelling:strict} efficiently - as we show in the next two subsections.

\subsubsection{Linear model}\label{sec:pp:exactsolutions:linearmodel}
We consider linear classifiers as defined in~\refeq{eq:linearmodel}.
We can rewrite~\refeq{eq:pertinentpositive:modelling:strict} as follows:
\begin{equation}\label{eq:pertinentpositive:modelling:strict:linearclassifier}
\begin{split}    
    &\underset{\setI}{\min}\,|\setI| \quad\quad \text{s.t. } \yorig b + \sum_{i\,\in\,\setI} z_i > 0
\end{split}
\end{equation}
where we defined
\begin{equation}\label{eq:pp:exactsolution:linearmodel:sum}
    z_i = (\w)_i(\xorig)_i\yorig
\end{equation}
The new optimization problem~\refeq{eq:pertinentpositive:modelling:strict:linearclassifier} reduces the original problem~\refeq{eq:pertinentpositive:modelling:strict} to a problem in which we want to find a minimal subset of real numbers $z_i$ such that their sum is strictly greater than a given constant ($\yorig b$). We can find such a subset of numbers (features) by sorting all $z_i$ in a descending order and then select the first $k$ such that $\yorig b + \sum_i^k z_i > 0$. Finally, the construction of $\setI$ follows immediately since each $z_i$ corresponds uniquely to the $i$-th feature.
Note that it might be the case that for some pairs $(\xorig, \yorig)$ no feasible solution exists -  in such a case we have to fall back to the proposed 2-phase algorithm (see previous section). Also note that the globally optimal set $\setI$ is not necessarily unique - however, the size $|\setI|$ is unique.

\subsubsection{Special quadratic model}\label{sec:pp:exactsolutions:quadraticmodel}
We consider quadratic classifiers~\refeq{eq:quadraticmodel} where the matrix $\mat{Q}$ is a diagonal matrix - e.g. Gaussian Naive Bayes classifier:
\begin{equation}
    \mat{Q} = \diag(\alpha_i) \quad \forall\,i: \alpha_i\in\RN
\end{equation}
Similar to the previous case of a linear classifier, we can rewrite the constraint~\refeq{eq:pertinentpositive:modelling:strict:constraint} as an independent of sum of weighted features - i.e. every summand uniquely corresponds to a feature:
\begin{equation}\label{eq:pp:exactsolution:quadraticmodel:sum}
    z_i = \alpha_i(\xorig)_i^2\yorig + (\q)_i(\xorig)_i\yorig
\end{equation}
Like in the case of the linear classifier, we can find a globally optimal solution by finding a minimal subset of indices $i$ such that $\yorig b + \sum_i z_i > 0$. Again, we can do so by sorting all $\z_i$ in descending order and select the first $k$ item such that $\yorig b + \sum_i^k z_i > 0$ - again, note that the global optimum is not necessarily unique and there might not even exist a feasible solution for every possible $\xorig$ and $\yorig$.

\subsection{Plausibility}\label{sec:pp:plausibility}
So far, we ignore the aspect of plausibility - i.e. making sure that the pertinent positive is realistic and plausible in the data domain.

Here we propose to make use of a known density based method for computing plausible counterfactual explanations~\cite{plausiblecounterfactuals}. The authors of~\cite{plausiblecounterfactuals} propose to use a Gaussian Mixture Model (GMM) for estimating the density from a given training set $\densityestimatorgmm(\x) = \sum_{j=1}^{m} \pi_j \N(\x \mid \vec{\mu}_j,\mat{\Sigma}_j)$.
Then they propose to use the following component wise approximation of $\densityestimatorgmm(\x)$ as an additional quadratic convex constraint in the optimization problem for computing plausible counterfactual explanations:
\begin{equation}\label{eq:gmm:componentwiseapprox}
    (\x - \vec{\mu}_j)^\top\mat{\Sigma}_j^{-1}(\x - \vec{\mu}_j) + c_j \leq {\densitythreshold}'
\end{equation}
where ${\densitythreshold}'$ denotes a density threshold that ensures that the solution $\x$ lies in a region of high density\footnote{The definition of the other constants in~\refeq{eq:gmm:componentwiseapprox} can be found in~\cite{plausiblecounterfactuals}.}. Since their proposed approximation is a component wise approximation, they get $m$ different constraints of the form~\refeq{eq:gmm:componentwiseapprox} and therefore have to solve the original optimization problem $m$ times (each time with a different constraint~\refeq{eq:gmm:componentwiseapprox}) - in the end they select the solution that minimizes the original objective (in their case the original objective is closeness to the original sample).
Note that we can simply transfer this approach of plausibility to the setting of computing plausible pertinent positives because the approach~\cite{plausiblecounterfactuals} is completely independent of the original objective and other constraints - i.e. it is a general approach of ensuring that the solution of a mathematical program lies in a region of high density.
The appealing benefit of using this approach is that we can guarantee that the resulting solution (pertinent positive) lies in a region of high density (i.e. it is plausible and realistic under the fitted GMM) although we give up closeness which we argue is not that important when it comes to plausibility.

In section~\ref{sec:pp:experiments} we empirically evaluate the quality of the plausible pertinent positives as computed by this approach and compare them with the closest pertinent positives as computed by our proposed 2-phase algorithm (see section~\ref{sec:pp:2phasealgo}).

\subsection{Experiments}\label{sec:pp:experiments}
\begin{table*}[tb]
    \centering
    \caption{Computation of contrastive explanations of different models on different standard benchmark data sets. We compute the sparsity \refeq{eq:score:sparsity}, the closeness to the original sample \refeq{eq:score:closeness} and the overlap of perturbed features in the pertinent negative vs. ``turned on'' features in the pertinent positive. We report the mean and variance (rounded to two decimal places) - for sparsity larger values are better whereas for closeness and the feature overlap smaller values are better. ``Closeness+'' denotes the closeness of the pertinent positive computed by the 2-phase algorithm (Algorithm \ref{algo:2phase}) and ``FeatOverlap'' denotes the overlap of ``turned on'' features in the pertinent positive and the perturbed features in the pertinent negative.}
    \label{tab:experiments:results}
    \begin{tabular}{cc|c|c|c}
        \hline
        Dataset & Scores & LogisticRegression & QDA & GLVQ \\
        \hline 
        \hline
        \multirow{4}{*}{\oneLine{Iris}}
        &  Sparsity & $3.0 (\pm 0.0)$ & $1.38 (\pm 0.3)$ & $3.0 (\pm 0.0)$ \\
        &  Closeness & $0.62 (\pm 0.06)$ & $2.01 (\pm 0.58)$ & $0.57 (\pm 0.14)$ \\
        &  Closeness+ & $0.01 (\pm 0.0)$ & $0.11 (\pm 0.09)$ & $0.14 (\pm 0.05)$ \\
        & FeatOverlap & $1.0 (\pm 0.0)$ & $2.62 (\pm 0.3)$ & $0.0 (\pm 0.0)$\\
        \cline{1-5}
        \hline
        
        \cline{1-5}
        \multirow{4}{*}{\twoLines{House}{prices}}
        &  Sparsity & $8.0 (\pm 0.0)$ & $5.68 (\pm 1.2)$ & $8.0 (\pm 0.0)$ \\
        &  Closeness & $0.53 (\pm 0.0)$ & $1.59 (\pm 0.36)$ & $0.93 (\pm 0.28)$ \\
        &  Closeness+ & $0.0 (\pm 0.0)$ & $0.43 (\pm 0.24)$ & $0.37 (\pm 0.21)$ \\
        & FeatOverlap & $1.0 (\pm 0.0)$ & $3.32 (\pm 1.2)$ & $1.0 (\pm 0.0)$\\
        \cline{1-5}
        \hline
        
        \cline{1-5}
        \multirow{4}{*}{\twoLines{Breast}{cancer}}
        &  Sparsity & $29.0 (\pm 0.0)$ & $17.38 (\pm 9.06)$ & $29.0 (\pm 0.0)$ \\
        &  Closeness & $0.89 (\pm 0.56)$ & $9.42 (\pm 35.94)$ & $2.59 (\pm 0.64)$ \\
        &  Closeness+  & $0.89 (\pm 0.56)$ & $0.61 (\pm 0.63)$ & $2.59 (\pm 0.66)$ \\
        & FeatOverlap & $1.0 (\pm 0.0)$ & $12.62 (\pm 9.06)$ & $0.04 (\pm 0.04)$\\
        \hline
        
        \cline{1-5}
        \multirow{4}{*}{\oneLine{Wine}}
        &  Sparsity & $12.0 (\pm 0.0)$ & $7.7 (\pm 3.44)$ & $11.45 (\pm 0.25)$ \\
        &  Closeness & $0.61 (\pm 0.07)$ & $3.54 (\pm 3.66)$ & $1.05 (\pm 0.31)$ \\
        &  Closeness+  & $0.0 (\pm 0.0)$ & $0.45 (\pm 0.42)$ & $0.05 (\pm 0.01)$ \\
        & FeatOverlap & $1.0 (\pm 0.0)$ & $5.3 (\pm 3.44)$ & $1.1 (\pm 0.99)$\\
        \hline
    \end{tabular}
\end{table*}
We want to empirically verify that our proposed modelling yields pertinent positives that fit the intuition of a pertinent positive as discussed in the introduction. We therefore evaluate our proposed modelling and the derived mathematical programs on a set of different standard benchmark data sets. We compare the results of \refeq{eq:opt:pertinentpositive} with those of the 2-phase algorithm Algorithm~\ref{algo:2phase}. Since the convex programs are guaranteed to output valid pertinent positive, we would have to validate the outputs (check if it is a valid pertinent positive) of the non-convex programs only (e.g. DCs for quadratic and LGMLVQ models) - however, we can neglect this in our specific situation because we choose a specific solver that is guaranteed to output a feasible solution.

\paragraph*{Evaluation measures}
For the quantitative evaluation of the computed pertinent positives, we choose two scoring functions for assessing sparsity and closeness to the original sample. We evaluate sparsity of a pertinent positive $\xcf$ with \refeq{eq:score:sparsity}\footnote{We compare the sparseness of the original data point with the sparseness of the pertinent positive.} and closeness to the original sample $\xorig$ with \refeq{eq:score:closeness}.
\begin{equation}\label{eq:score:sparsity}
\pnorm{\xorig}_{0} - \pnorm{\xcf}_{0}
\end{equation}
\begin{equation}\label{eq:score:closeness}
\sum_{i\,\in\,\setI}\big|(\xcf)_i - (\xorig)_i\big| 
\end{equation}

\paragraph*{Experimental setup}
We run the experiments on four standard benchmark sets using logistic regression, a quadratic discriminant analysis (QDA) and GLVQ. We use the "Iris Plants Data Set"~\cite{irisdata}, the "Wine data set"~\cite{winedata}, the "Ames Housing dataset"~\cite{housepricedata}\footnote{We turn it into a binary classification problem by setting the target to $1$ if the price is greater or equal to 160k\$ and $0$ otherwise. In addition, we select the following features: TotalBsmt, 1stFlr, 2ndFlr, GrLivA, WoodDeck, OpenP, 3SsnP, ScreenP and PoolA} and the "Breast Cancer Wisconsin (Diagnostic) Data Set"~\cite{breastcancer}.
We compute a three-fold cross validation and compute a pertinent positive by only solving \refeq{eq:opt:pertinentpositive} and another one by using our proposed 2-phase algorithm (Algorithm~\ref{algo:2phase}). We standardize all data sets, use a regularization strength of $1.0$ when estimating the covariance matrices in QDA, set the basis values to $\basisvalues=\vec{0}$ and set the threshold for ``turned on'' features \refeq{eq:setI} to $\epsilon=0$ for all data sets. We report the mean sparsity \refeq{eq:score:sparsity} and the mean closeness \refeq{eq:score:closeness} for each combination of model, method and data set (we also report the variance) - because the sparsity does not change when using the 2-phase algorithm instead of \refeq{eq:opt:pertinentpositive} only, we only report sparsity once.
For the purpose of better observing the properties of non-trivial pertinent positives, we always exclude the class of the zero vector - as discussed in section ~\ref{sec:pp:2phasealgo:closeness}, all samples from the class $\classifier(\vec{0})$ would yield the sparsest and trivial pertinent positive $\vec{0}$ which makes them less suited for evaluating our proposed algorithms.
In case of logistic regression, we also compute a globally optimal solution of a strict pertinent positive as modelled in section~\ref{sec:pp:exactsolutions}. We compare the sparsity of these strict pertinent positives with the sparsity of the pertinent positives as computed by our proposed 2-phase algorithm.
In addition, we compare the feature overlap between pertinent negatives and pertinent positives. 

For the purpose of informative and useful explanations it is beneficial that the pertinent positive and the pertinent negative ``share'' as few features as possible - meaning that the overlap of ``turned on'' features in the pertinent positive and the perturbed features in a pertinent negative should be rather small. We argue that if the pertinent positive and the pertinent negative ``share'' many features they might not be that useful and informative\ifthenelse{\boolean{arxiv}}{\footnote{This depends of course a lot on the specific situation and use case.}}{} - if the overlap of features happens to be too large, one could add additional constraints to the optimization problems for manually including or excluding some features that finally result in a smaller overlap of features.

We compute the pertinent negatives by using a Python toolbox~\cite{ceml} for efficiently computing counterfactual explanations - we use the  $1$-norm as a regularization for enforcing sparsity.
We also keep track of the F1-score to ensure that the classifiers learned a ``somewhat reasonable'' decision boundary - because all classifiers perform quite well, we do not report the  F1-scores in here and refer the interested reader to the published source code and protocols. We approximately solve the non-convex QPs using the convex-concave penalty (CCP) method~\cite{qcqp}. Because the CCP method is guaranteed to output a feasible solution, we do not have to check if the pertinent positive is valid. Further details (including the raw protocols of the experiments) and the implementations itself is available on GitHub\footnote{\url{https://github.com/andreArtelt/contrastive_explanations}}.
The results are shown in Table~\ref{tab:experiments:results} whereby more details can be found in the raw protocols of the experiments that are available on GitHub.

\paragraph*{Discussion of results}
We observe that our proposed method is able to consistently compute sparse pertinent positives. Furthermore, we observe that our proposed 2-phase algorithm significantly increases the closeness of the pertinent positives to the original samples. Only in the case of GLVQ and logistic regression in combination with the breast cancer data set, the 2-phase algorithm is not able to improve on average upon \refeq{eq:opt:pertinentpositive} - we think that this might be an issue of unfavorable chosen hyperparameters\footnote{Note that we use the same hyperparameters over all data sets.} (we expect that changing the model would most likely show a difference). In addition, the large variances in the results of QDA for the breast cancer data set can be explained by some outliers.
Also note that the mean sparsity is often just a little bit below the total number of features. This means that our proposed method was able to ``turn off'' many features which perfectly fits the intuition of a pertinent positive as discussed in the introduction.
In case of logistic regression, we observe that the sparsity of the (globally optimal) strict pertinent positives with the sparsity of the pertinent positives as computed by our proposed 2-phase algorithm are equal for all data sets. We argue that this is a strong indicator for the powerfullness of our proposed 2-phase algorithm since it can reproduce a globally optimal strict pertinent positive although it is an approximation only.
Finally, we observe that the overlap of ``turned on'' features in the pertinent positives and the perturbed features in the pertinent negatives is relatively small. This means that the pertinent positives and the pertinent negatives ``share'' only very few features in their explanations which makes them useful and informative in practice - as discussed previously, if both explanations would use (more or less) the same features, they would not be that informative to a user. However, please note that these findings are empirically only and might not necessarily generalize to other models and/or data sets.

\paragraph*{Evaluating plausibility constraints}
In order to demonstrate the effectiveness of the plausibility constraints (see section~\ref{sec:pp:plausibility}), we compute and compare sparsest pertinent positives (as computed by our proposed 2-phase algorithm) and (approximately) closest plausible pertinent positives of the "Optical Recognition of Handwritten Digits Data Set"~\cite{ocr} under a softmax regression model - details on the hyper parameters can be found in the source code.
In addition, we also compute a closest pertinent negative (we always choose $\ycf=0$ as a target label) and compare it with the closest pertinent positive of the original class.

\begin{figure*}[!htb]
  \caption{Samples from the digit data set. \textit{First block:} Original samples. \textit{Second block:} Closest pertinent negative where we always chose the same target label $0$. \textit{Third block:} Pertinent positives generated without any density/plausibility constraints. \textit{Fourth block:} Pertinent positives generated with the proposed plausibility constraint. The corresponding labels are shown below each image.}
  \label{fig:digits:samples}

  \centering
  \vspace*{0.5cm}
  Original samples  
  
  \begin{minipage}[b]{0.19\textwidth}
    \includegraphics[width=\textwidth]{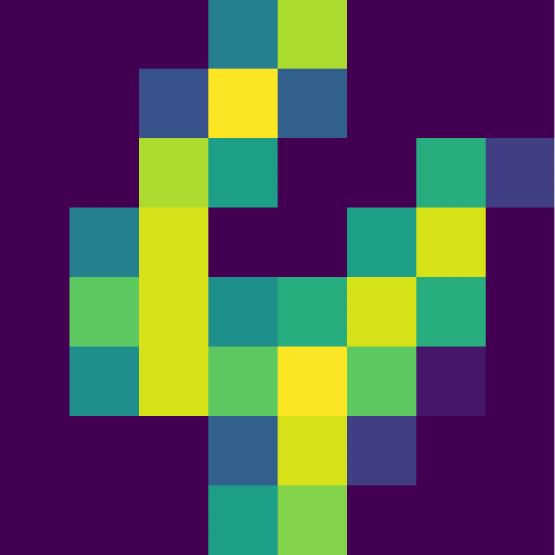}  
    \caption{Label: 4} 
   \end{minipage}
  \hfill
  \begin{minipage}[b]{0.19\textwidth}
    \includegraphics[width=\textwidth]{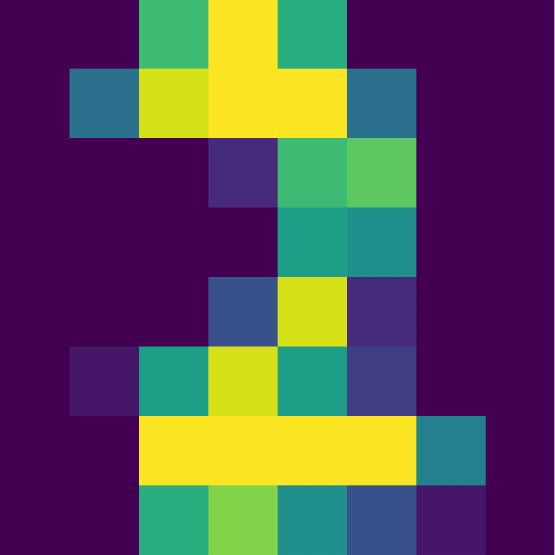}
    \caption{Label: 2} 
  \end{minipage}
  \hfill
  \begin{minipage}[b]{0.19\textwidth}
    \includegraphics[width=\textwidth]{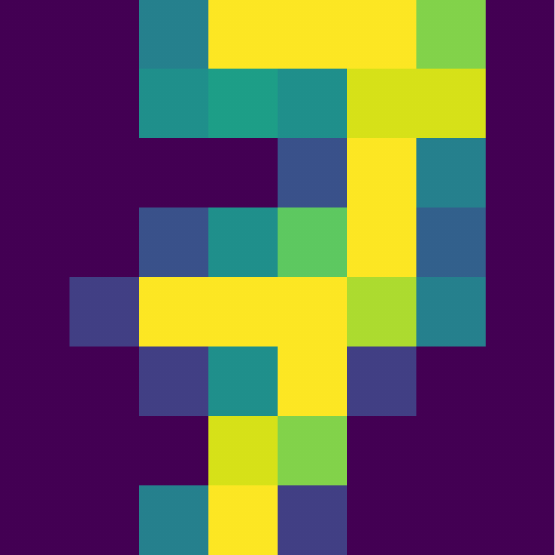}
    \caption{Label: 7} 
  \end{minipage}
  \hfill  
  \begin{minipage}[b]{0.19\textwidth}
    \includegraphics[width=\textwidth]{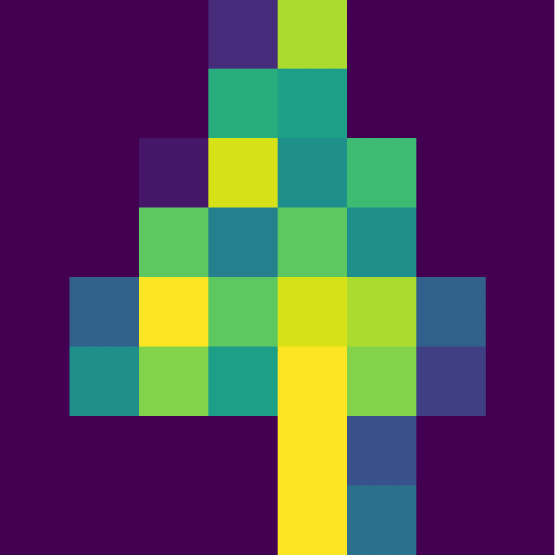}
    \caption{Label: 4} 
  \end{minipage}
  \hfill  
  \begin{minipage}[b]{0.19\textwidth}
    \includegraphics[width=\textwidth]{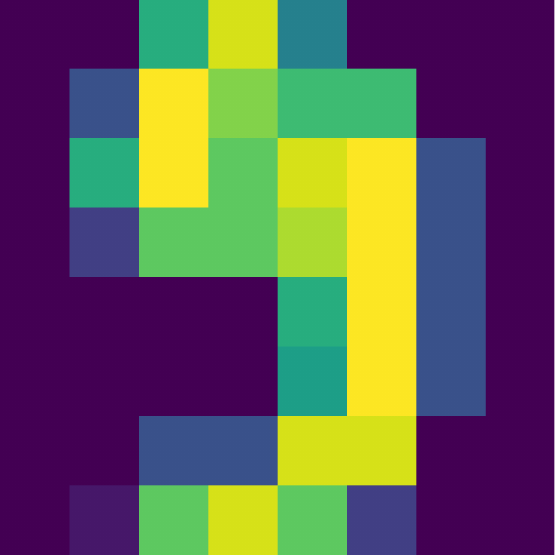}
    \caption{Label: 9} 
  \end{minipage}  
 
  \vspace*{0.1cm}
  \rule[2ex]{18cm}{2.0pt}  
  Closest \textit{pertinent negative} under a \underline{softmax regression model}
  \vfill
 \begin{minipage}[b]{0.19\textwidth}
    \includegraphics[width=\textwidth]{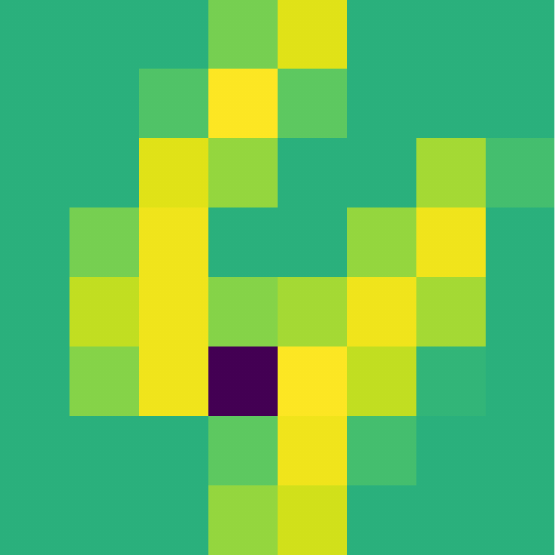}
    \caption{Label: 0} 
  \end{minipage}
  \hfill
  \begin{minipage}[b]{0.19\textwidth}
    \includegraphics[width=\textwidth]{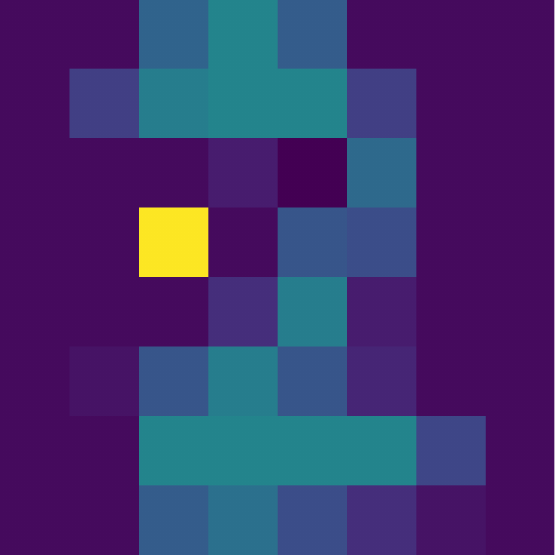}
    \caption{Label: 0} 
  \end{minipage}
  \hfill
  \begin{minipage}[b]{0.19\textwidth}
    \includegraphics[width=\textwidth]{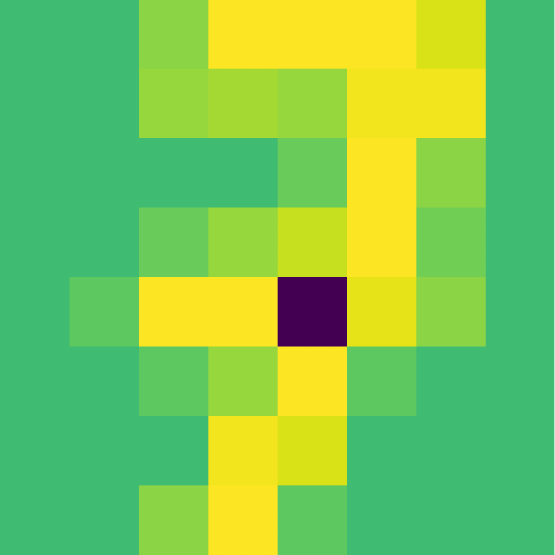}
    \caption{Label: 0} 
  \end{minipage}
  \hfill
  \begin{minipage}[b]{0.19\textwidth}
    \includegraphics[width=\textwidth]{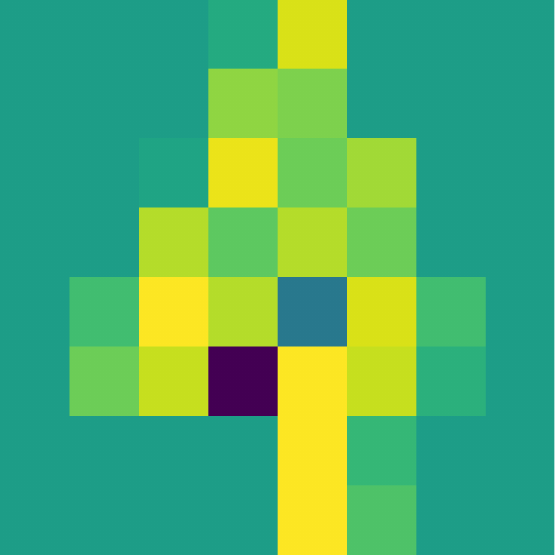}
    \caption{Label: 0} 
  \end{minipage}
  \hfill
  \begin{minipage}[b]{0.19\textwidth}
    \includegraphics[width=\textwidth]{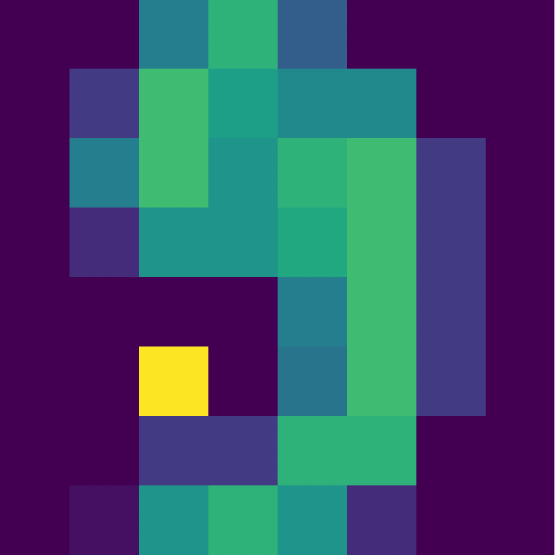}
    \caption{Label: 0} 
  \end{minipage}
 
  \vspace*{0.1cm}
  \rule[2ex]{18cm}{2.0pt}  
  Closest \textit{pertinent positives} under a \underline{softmax regression model}
  \vfill
  \begin{minipage}[b]{0.19\textwidth}
    \includegraphics[width=\textwidth]{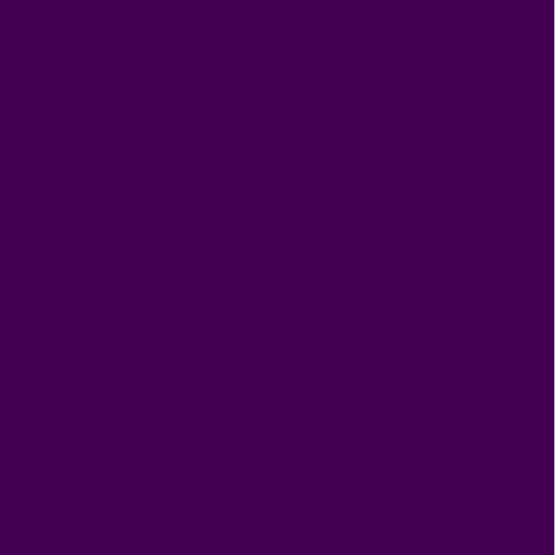}
    \caption{Label: 4} 
  \end{minipage}
  \hfill
  \begin{minipage}[b]{0.19\textwidth}
    \includegraphics[width=\textwidth]{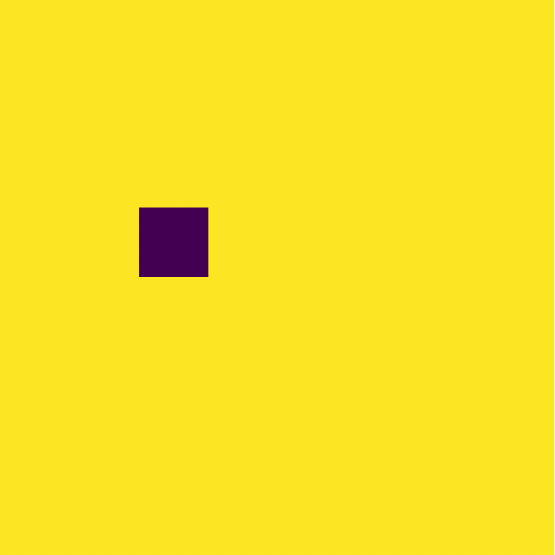}
    \caption{Label: 2} 
  \end{minipage}
  \hfill
  \begin{minipage}[b]{0.19\textwidth}
    \includegraphics[width=\textwidth]{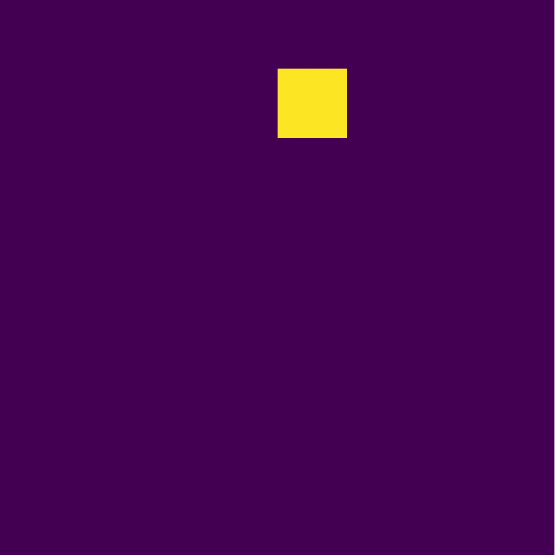}
    \caption{Label: 7} 
  \end{minipage}
  \hfill
  \begin{minipage}[b]{0.19\textwidth}
    \includegraphics[width=\textwidth]{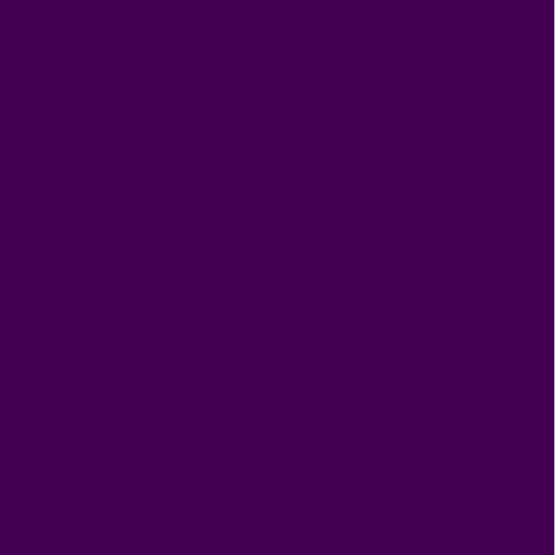}
    \caption{Label: 4} 
  \end{minipage}
  \hfill
  \begin{minipage}[b]{0.19\textwidth}
    \includegraphics[width=\textwidth]{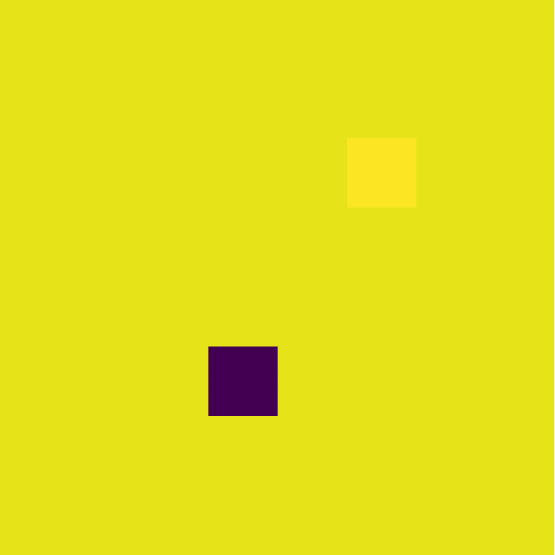}
    \caption{Label: 9} 
  \end{minipage}
  
  \vspace*{0.1cm}
  \rule[2ex]{18cm}{2.0pt}   
  Closest \textit{plausible pertinent positive} under a \underline{softmax regression model}  
   \vfill
  \begin{minipage}[b]{0.19\textwidth}
    \includegraphics[width=\textwidth]{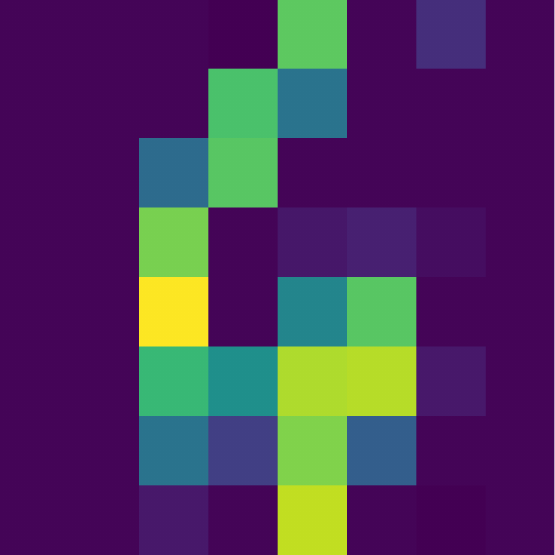}
    \caption{Label: 4} 
  \end{minipage}
  \hfill
  \begin{minipage}[b]{0.19\textwidth}
    \includegraphics[width=\textwidth]{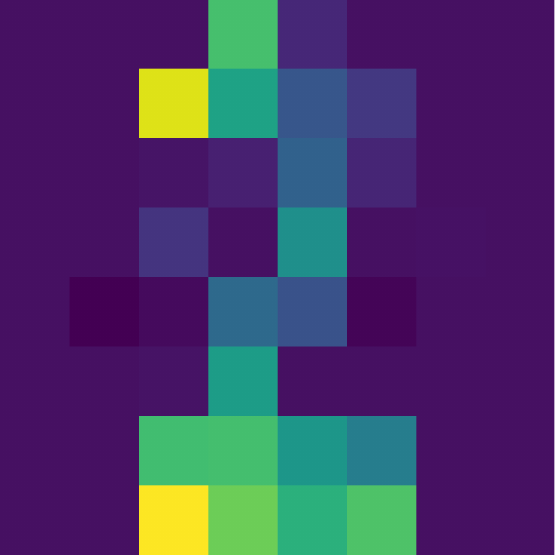}
    \caption{Label: 2} 
  \end{minipage}
  \hfill
  \begin{minipage}[b]{0.19\textwidth}
    \includegraphics[width=\textwidth]{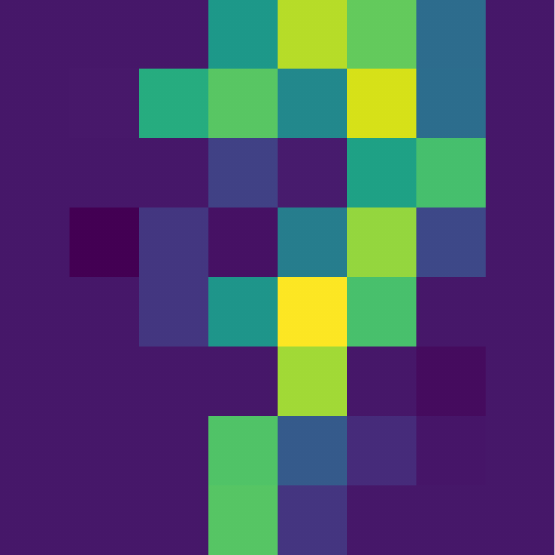}
    \caption{Label: 7} 
  \end{minipage}
  \hfill
  \begin{minipage}[b]{0.19\textwidth}
    \includegraphics[width=\textwidth]{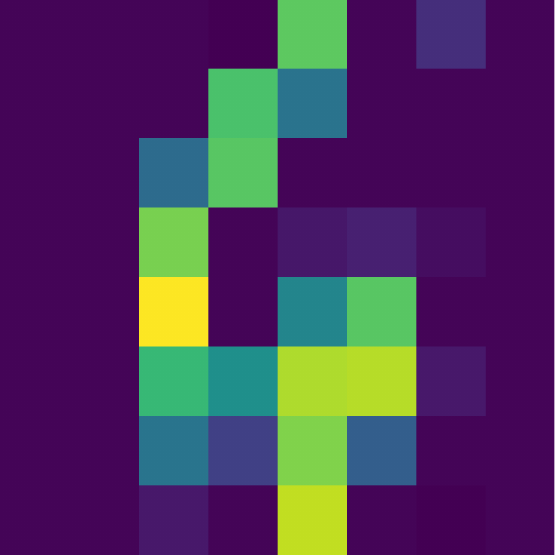}
    \caption{Label: 4} 
  \end{minipage}
  \hfill
  \begin{minipage}[b]{0.19\textwidth}
    \includegraphics[width=\textwidth]{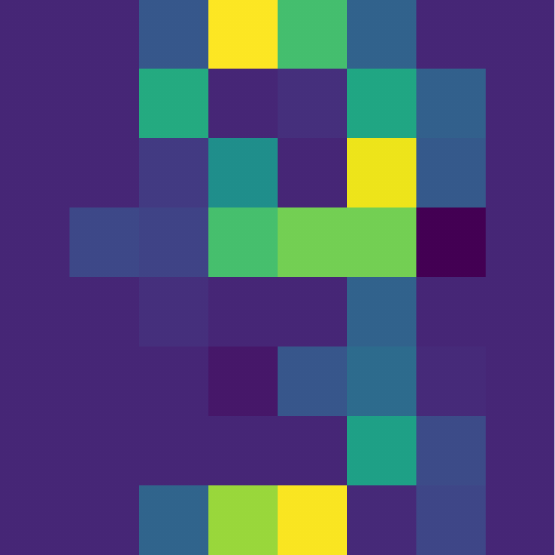}
    \caption{Label: 9}
  \end{minipage}
\end{figure*}

The results are shown in~\reffig{fig:digits:samples}. We observe that the sparsest pertinent positives are by no way plausible (often very few features or no features at all are already sufficient for the prediction) but the closest pertinent positives under plausibility constraints are plausible. We also observe that a closest pertinent negative and a sparsest pertinent positive are quite different from each other - which perfectly agrees with our observations of a low feature overlap in Table~\ref{tab:experiments:results}. 

\section{Discussion and Conclusion}
In this work we extensively studied the computation of contrastive explanations that consists of a pertinent negative and a pertinent positive. We argued that computing a pertinent negative is equivalent to computing a counterfactual explanation - this reduction enables us to use methods from the counterfactual explanations literature for efficiently computing pertinent negatives. We also proposed to model pertinent positives as a constrained optimization problem and proposed upon that a 2-phase algorithm for computing  qualitatively better pertinent positives. Building upon these, we derived mathematical programs for efficiently computing pertinent positives of many standard ML models. Furthermore, we proposed a stricter modelling of pertinent positives that allowed us to exactly efficiently compute pertinent positives. We also successfully applied ideas for computing plausible counterfactual explanations to the problem of computing plausible pertinent positives. Finally, we empirically evaluated our proposed methods on several standard benchmark data sets.

\ifthenelse{\boolean{arxiv}}{
\appendix

\section{Proofs and Derivations}\label{sec:appendix}
\subsection{Pertinent positives of linear models}\label{appendix:pp:linearmodels}
We assume $\sety=\{-1,1\}$ and $\classifier(\x)=\sign\left(\w^\top\x + b\right)$. We then can rewrite the constraint \refeq{eq:opt:pertinentpositive:constraint} as follows:
\begin{equation}
\begin{split}
& \classifier(\xorig - \deltacf) = \yorig \\
& \Leftrightarrow \yorig\Big(\classifier(\xorig - \deltacf)\Big) > 0 \\
& \Leftrightarrow \yorig\Big(\w^\top\left(\xorig - \deltacf\right) + b\Big) > 0 \\
& \Leftrightarrow \yorig\Big(\w^\top\xorig - \w^\top\deltacf + b\Big) > 0 \\
& \Leftrightarrow \underbrace{\yorig\w^\top\xorig + \yorig b}_{\text{constant } := c} - \yorig\w^\top\deltacf > 0 \\
& \Leftrightarrow \yorig\w^\top\deltacf - c < 0
\end{split}
\end{equation}
where we temporarily ignored the special case of $\sign(0)$.

Finally, we relax the strict inequality by adding a small positive number $\epsilon$ to the left side - by doing this we avoid that the resulting data points lies on the decision boundary (in this case the $\sign$ would be undefined):
\begin{equation}\label{eq:appendix:pp:linearmodel:constraint:final}
\yorig\w^\top\deltacf - c + \epsilon \leq 0
\end{equation}
Note that \refeq{eq:appendix:pp:linearmodel:constraint:final} is linear in $\deltacf$ - thus the final optimization problems become linear programs (LPs) which can be solved very efficiently~\cite{Boyd2004}.

In case of a multi-class problem, we would get multiple constraints of the form \refeq{eq:appendix:pp:linearmodel:constraint:final} - however, since they are all linear, the final problems are still LPs.

\subsection{Pertinent positives of quadratic models}\label{appendix:pp:quadraticmodels}
We assume $\sety=\{-1,1\}$ and $\classifier(\x)=\sign\left(\x^\top\mat{Q}\x + \vec{q}^\top\x +  c\right)$ with $\mat{Q}\in\SetSymMat^{\dimsym}$. We then can rewrite the constraint \refeq{eq:opt:pertinentpositive:constraint} as follows:
\begin{equation}
\begin{split}
& \classifier(\xorig - \deltacf) = \yorig \\
& \Leftrightarrow \yorig\,\classifier(\xorig - \deltacf) > 0 \\
& \Leftrightarrow \yorig\Big(\left(\xorig - \deltacf\right)^\top\mat{Q}\left(\xorig - \deltacf\right) + \vec{q}^\top\left(\xorig - \deltacf\right) + c\Big) > 0 \\
& \Leftrightarrow \yorig\Big(\xorig^\top\mat{Q}\xorig - \xorig^\top\mat{Q}\deltacf -\\
& \quad\quad\quad\quad \deltacf^\top\mat{Q}\xorig + \deltacf^\top\mat{W}\deltacf + \vec{q}^\top\xorig - \vec{q}^\top\deltacf + c \Big) > 0 \\
& \Leftrightarrow \underbrace{\yorig\Big(\xorig^\top\mat{Q}\xorig + \vec{q}^\top\xorig + c\Big)}_{\text{constant } := -c'} + \underbrace{-2\yorig\xorig^\top\mat{Q}}_{\text{constant } := -\z^\top}\deltacf +\\
&\quad\quad\yorig\deltacf^\top\mat{Q}\deltacf > 0\\
& \Leftrightarrow \deltacf^\top\mat{\tilde{Q}}\deltacf + \deltacf^\top\z + c' < 0
\end{split}
\end{equation}
where we defined
\begin{equation}
\mat{\tilde{Q}} = -\yorig\mat{Q}
\end{equation}
Again, we relax the strict inequality by adding a small positive number $\epsilon$ to the left side:
\begin{equation}\label{eq:appendix:pp:quadraticmodel:constraint}
\deltacf^\top\mat{\tilde{Q}}\deltacf + \deltacf^\top\z + c' + \epsilon \leq 0
\end{equation}
A basic fact from linear algebra states that we can rewrite every real symmetric matrix as the difference of two s.psd. matrices.  
Furthermore, in case of QDA or Gaussian Naive Bayes such a decomposition appears naturally because in both cases the matrix $\mat{Q}$ is defined as the difference of two (s.psd.) covariance matrices.
Assuming that we decompose $\mat{\tilde{Q}}$ as
\begin{equation}
\mat{\tilde{Q}} = \mat{\tilde{Q}}_1 - \mat{\tilde{Q}}_2 \quad\quad \mat{\tilde{Q}}_1,\mat{\tilde{Q}}_2 \in{\SetSymMat}_{+}^{\dimsym} 
\end{equation}
we can rewrite \refeq{eq:appendix:pp:quadraticmodel:constraint} as follows:
\begin{equation}\label{eq:appendix:pp:quadraticmodel:constraint:dc}
\underbrace{\deltacf^\top\mat{\tilde{Q}}_1\deltacf + \deltacf^\top\z + c' + \epsilon}_{\text{convex in } \deltacf} - \underbrace{\deltacf^\top\mat{\tilde{Q}}_2\deltacf}_{\text{convex in } \deltacf} \leq 0
\end{equation}
Clearly, \refeq{eq:appendix:pp:quadraticmodel:constraint:dc} is now a difference of convex quadratic functions which turns the resulting optimization problem into a DC for which good approximatation solvers like the Suggest-and-Improve framework exist~\cite{qcqp}.

In case of a multi-class problem, we would get multiple constraints of the form \refeq{eq:appendix:pp:quadraticmodel:constraint:dc} - however, since they are all of the same form, the final optimization problems are still DCs.

\subsection{Pertinent positives of LVQ models}\label{appendix:pp:lvqmodels}
If the data point $\xorig - \deltacf$ is classified as $\yorig$, we know that the closest prototype must be one labeled as $\yorig$. Therefore, for each suitable prototype $\vec{\prototype}_i)$ (that is $\protolabel_i=\yorig$), we get the following set of constraints:
\begin{equation}
\dist(\xorig - \deltacf, \vec{\prototype}_i) < \dist(\xorig - \deltacf, \vec{\prototype}_j) \quad \forall\,j: \,\protolabel_j\neq\yorig
\end{equation}
After rearranging the terms and relaxing the strict inequality by adding a small positive $\epsilon$, we get:
\begin{equation}\label{eq:appendix:pp:lvq:constraint:general}
\dist(\xorig - \deltacf, \vec{\prototype}_i) - \dist(\xorig - \deltacf, \vec{\prototype}_j) + \epsilon \leq 0 \quad \forall\,j: \,\protolabel_j\neq\yorig
\end{equation}
Fixing $i$ and $j$ and plugging the most general distance function $\dist(\x, \vec{\prototype}_j) = (\x - \vec{\prototype}_j)^\top{\distmat}_{j}(\x - \vec{\prototype}_j)$ with $\distmat_j\in\SetSymMat_{+}^{d}$ (LGMLVQ) into \refeq{eq:appendix:pp:lvq:constraint:general}, yields:
\begin{equation}\label{eq:appendix:pp:lvq:constraint:final}
\begin{split}
& \dist(\xorig - \deltacf, \vec{\prototype}_i) - \dist(\xorig - \deltacf, \vec{\prototype}_j) + \epsilon \leq 0 \\
& \Leftrightarrow \left(\xorig - \deltacf - \vec{\prototype}_i\right)^\top{\distmat}_i\left(\xorig - \deltacf - \vec{\prototype}_i\right) -\\
& \quad\quad \left(\xorig - \deltacf - \vec{\prototype}_j\right)^\top{\distmat}_j\left(\xorig - \deltacf - \vec{\prototype}_j\right) + \epsilon \leq 0 \\
& \Leftrightarrow \left(\xorig - \vec{\prototype}_i\right)^\top{\distmat}_i\left(\xorig - \vec{\prototype}_i\right) - 2\left(\xorig - \vec{\prototype}_i\right)^\top{\distmat}_i\deltacf + \\
&\quad\quad \deltacf^\top{\distmat}_i\deltacf - \left(\xorig - \vec{\prototype}_j\right)^\top{\distmat}_j\left(\xorig - \vec{\prototype}_j\right) + \\
&\quad\quad 2\left(\xorig - \vec{\prototype}_j\right)^\top{\distmat}_j\deltacf - \deltacf^\top{\distmat}_j\deltacf + \epsilon \leq 0 \\
& \Leftrightarrow \deltacf^\top\left(\underbrace{{\distmat}_i - {\distmat}_j}_{:= \mat{A}_{ij}}\right)\deltacf +\deltacf^\top\underbrace{\Big(2{\distmat}_j \left(\xorig- \vec{\prototype}_j\right) - 2{\distmat}_i
\left(\xorig- \vec{\prototype}_i\right)\Big)}_{\text{constant } := \q_{ij}} \\
& \quad +\underbrace{\left(\xorig - \vec{\prototype}_i\right)^\top{\distmat}_i\left(\xorig - \vec{\prototype}_i\right) - \left(\xorig - \vec{\prototype}_j\right)^\top{\distmat}_j\left(\xorig - \vec{\prototype}_j\right)}_{\text{constant } := c_{ij}} \\&\quad+ \epsilon \leq 0
\end{split}
\end{equation}
Because all we can say about $\mat{A}_{ij}$ is that it is symmetric, the constraints \refeq{eq:appendix:pp:lvq:constraint:final} are quadratically non-convex. However, like we did in case of quadratic models (see appendix~\ref{appendix:pp:quadraticmodels}), we can rewrite the constraints \refeq{eq:appendix:pp:lvq:constraint:final} as a difference of convex functions\footnote{In fact \refeq{eq:appendix:pp:lvq:constraint:final} decomposes naturally into a difference of convex functions because the only non-convex part $\mat{A}_{ij}$ is equal to the difference of two s.psd. distance matrices.} and turn the whole optimization problem into a special case of a DC. Therefore, in case of LGMLVQ, the optimization problems are non-convex and can only be approximately solved (e.g. via a DC).

In case of G(M)LVQ, the distance matrices ${\distmat}_i$ are always the same. Thus, the constraint \refeq{eq:opt:pertinentpositive:constraint} becomes a set of linear constraints:
\begin{equation}\label{eq:appendix:pp:lvq:constraint:glvq}
\deltacf^\top\q_{ij} + c_{ij} + \epsilon \leq 0 \quad \forall\,j: \,\protolabel_j\neq\yorig
\end{equation}
As a consequence, the resulting optimization problems become LPs which can be solved very efficiently~\cite{Boyd2004}.
}{}

\bibliographystyle{IEEEtran}
\bibliography{bibliography.bib}

\end{document}